\documentclass[a4paper,fleqn]{cas-dc}
\usepackage[numbers, sort&compress]{natbib}

\def\tsc#1{\csdef{#1}{\textsc{\lowercase{#1}}\xspace}}
\tsc{WGM}
\tsc{QE}

\begin{document}

\let\WriteBookmarks\relax
\def\floatpagepagefraction{1}
\def\textpagefraction{.001}

\shorttitle{\rmfamily UMSPU: Universal Multi-Size Phase Unwrapping via Mutual Self-Distillation and Adaptive Boosting Ensemble Segmenters}    

\shortauthors{Lintong Du et al.}  

\title [mode = title]{UMSPU: Universal Multi-Size Phase Unwrapping via Mutual Self-Distillation and Adaptive Boosting Ensemble Segmenters}  


%

\author[add1]{Lintong Du  }
\ead{dulintong@sjtu.edu.cn}
\author[add1]{Huazhen Liu }
\author[add1]{Yijia Zhang}
\author[add1]{Shuxin Liu}
\author[add1]{Rongjun Shao}
\author[add1,add2,add3]{Yuan Qu}
\author[add1,add2,add3]{Jiamiao Yang \corref{cor1}}
\ead{jiamiaoyang@sjtu.edu.cn}

\cortext[cor1]{Corresponding author}

\affiliation[add1]{organization={the School of Electronic Information and Electrical Engineering},
	addressline={Shanghai Jiao Tong University}, 
	city={Shanghai},
	postcode={200240}, 
	country={China}}

\affiliation[add2]{organization={the Institute of Marine Equipment},
	addressline={Shanghai Jiao Tong University}, 
	city={Shanghai},
	postcode={200240}, 
	country={China}}

 \affiliation[add3]{organization={the Institute of Medical Robotics},
	addressline={Shanghai Jiao Tong University}, 
	city={Shanghai},
	postcode={200240}, 
	country={China}}

\begin{abstract}
Phase unwrapping is a crucial technique in phase measurement. Deep learning - based methods are widely studied due to their better noise resistance and speed. However, existing phase unwrapping networks are constrained by the receptive field range and sparse semantic information, unable to effectively process high-resolution images, which severely limits their application in practical scenarios. To address this issue, we propose a Mutual Self-Distillation (MSD) mechanism and an adaptive-boosting ensemble segmenter to construct a Universal Multi-Size Phase Unwrapping network (UMSPU). MSD realizes cross-layer supervised learning by optimizing the bidirectional Kullback - Leibler divergence of attention maps, ensuring the precise extraction of fine-grained semantic features across different resolutions. The adaptive boosting ensemble segmenter combines weak segmenters with different receptive fields into a strong segmenter, ensuring stable segmentation at different spatial frequencies.
The proposed mechanisms help UMSPU break the resolution limitations of previous networks, increasing the applicable resolution range from 256×256 to 2048×2048 (a 64-fold increase). It also enables the network to achieve highly robust and strongly generalized phase unwrapping effects with a lightweight architecture, taking only 22.66 ms to process a high-resolution image. This truly propels the deep learning-based phase unwrapping method from the scientific research level to the practical application level.\nocite{*}
\end{abstract}


\begin{highlights}
\item A Universal Multi-Size Phase Unwrapping (UMSPU) network is proposed, breaking the resolution limit of deep learning phase unwrapping methods, achieving a 64 - fold increase in the applicable resolution range.
\item The proposed mechanisms help a lightweight architecture to achieve high - resolution and highly generalized phase unwrapping, with the unwrapping time for a single high - resolution image being only 22.66 ms, pushing deep learning - based phase unwrapping from the scientific research level to the practical application level.
\item A mutual self-distillation mechanism is proposed to perform bidirectional attention distillation on the encoder and decoder, enabling accurate semantic information extraction under cross-resolution inputs.
\item An ensemble segmenter architecture is proposed, which ensembles three segmenters with different receptive fields through an adaptive boosting strategy to achieve stable segmentation of semantic features with different spatial frequencies.
\item A curl loss is proposed, which can ensure that the gradient field output by the network meets the physical constraint of irrotationality.

\end{highlights}



\maketitle

\section{Introduction}\label{}
Phase unwrapping is a crucial technique in fields such as structured light imaging \cite {structuredlight,structuredlight2,structuredlight3}, optical interferometry (OI) \cite {OI,OI2}, synthetic aperture radar \cite {InSAR,InSAR2}, and magnetic resonance imaging \cite {MRI}. In these fields, the phase contains important information, such as height information \cite {structlightheight, SARheight}, the uniformity of the magnetic field, and physiological details \cite {MRIinfor}. However, during the actual measurement process, the phase obtained through the phase shift method is usually confined within the range of 
 $(-\pi, \pi ]$, which is called the wrapped phase\cite {U2}. Phase unwrapping refers to the process of recovering the true phase from the wrapped phase. This is an ill-posed problem because multiple unwrapped phases may correspond to the same wrapped phase. Ideally, phase unwrapping can be achieved by adding an appropriate integer multiple of 2$\pi$ to each pixel according to the phase difference between adjacent pixels. This relies on the phase continuity assumption (Itoh condition) \cite{Itohcondition}, but in actual measurements, both noise and drastic phase changes will violate the Itoh condition. Therefore, how to recover the accurate phase in such an ill-posed problem is an important and challenging task.

 Conventional phase unwrapping methods include the path-following method \cite{pathfollowing1,pathfollowing2,branchcut,qualitymap} and the optimization method \cite{optimmethod1, optimmethod2,optimmethod3,optimmethod4,optimmethod5}. The path-following method seeks the optimal integration path and performs phase unwrapping along this path. However, this method suffers from the problem of error accumulation along the path \cite{wang2022deep}. The optimization method obtains the unwrapped phase by minimizing the difference between the unwrapped phase gradient and the wrapped phase gradient. This method usually yields relatively smooth results. Nevertheless, the presence of noise will lead to errors in the wrapped phase gradient, leading to deviations in the “smooth” results of this method \cite{13}. Moreover, these methods are slow in speed and thus difficult to meet the requirements of rapid measurement scenarios.

In recent years, due to the excellent performance of neural networks, they have become the mainstream research direction in this field and have brought about many breakthroughs. Deep learning-based phase unwrapping methods are classified into regression methods \cite{regression1,regression2,regression3,regression4,regression5}, wrap count methods \cite{count1,count2,count3,count4}, and wrap count gradient methods \cite{gradient1,gradient2,gradient3}.
\begin{figure*}
    \centering
    \includegraphics[width = 1.8\columnwidth]{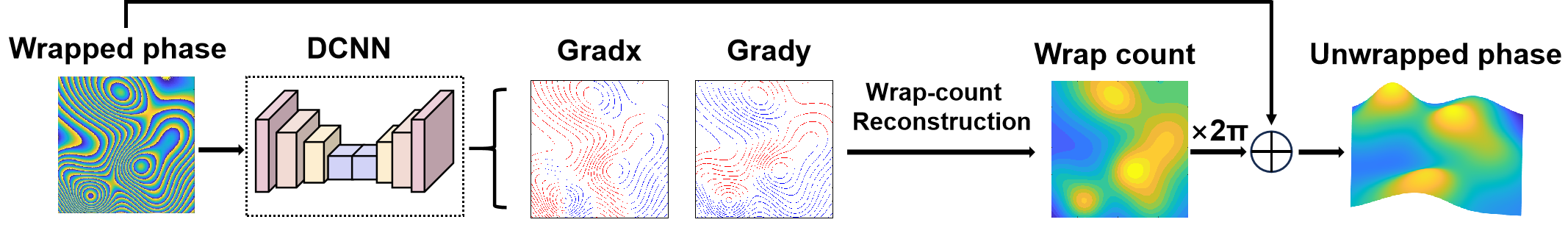}
    
    \caption{\rmfamily In the wrap count gradient method, the neural network classifies the wrap count gradient (0, +1, -1) of each point. After that, the wrap count is restored through the least squares method. Then, multiplying by 2$\pi$ and adding it to the wrapped phase yields the unwrapped phase. }
    \label{fig:gradient method}
     \vspace{-10pt} 
\end{figure*}
Jin et al. first proposed a neural network-based phase unwrapping method. They used convolutional neural networks (CNNs) to directly learn the mapping relationship from the wrapped phase map to the unwrapped phase map in the dataset. This method is called the regression method. Results showed that this method significantly outperformed conventional algorithms in both speed and accuracy \cite{jin2017deep}. Yair Rivenson et al. applied CNNs for holographic phase unwrapping, significantly accelerating holographic image reconstruction \cite{regression1}.  However, the regression method often fails to satisfy Eq. (\ref{phase}), introducing errors. To address this, G. E. Spoorthi et al. first reformulated phase unwrapping as a semantic segmentation problem and introduced PhaseNet, which classifies the wrap count for each pixel to achieve phase unwrapping \cite{count2}. This method is called the wrap count method. PhaseNet2.0 was later proposed. It was an improvement over PhaseNet, with enhanced loss functions and model structures, achieving better accuracy and noise resistance \cite{count3}. The wrap count method is fast and satisfies the constraints of Eq. (\ref{phase}). However, since it requires the classification of phase cycles, when the number of fringe cycles is dense, the accuracy and stability of the network will significantly decline. L. Zhou et al. proposed PGNet, which classifies the wrap count gradient at each point and uses an optimization algorithm to recover the unwrapped phase \cite{gradient1}. This method is called the wrap count gradient method. It avoids the direct classification of phase cycles. Instead, as shown in Fig. \ref{fig:gradient method}, it only requires the classification of the wrap count gradient, thus achieving better stability under dense fringes.

Previous research shows progress in various aspects of phase unwrapping. However, due to factors such as the receptive field of the model and the sparsity of gradient features, existing models are only applicable to images with a resolution of 256×256 or lower. This is inconsistent with the high-resolution images used in current optical measurements. Therefore, the limitation of resolution directly restricts the accuracy and practicality of deep learning phase unwrapping methods. Currently, there is a lack of research on expanding the applicable resolution of the phase unwrapping network.

In order to expand the applicable resolution range of deep learning phase unwrapping methods, we propose a universal multi-size phase unwrapping network (UMSPU) based on Mutual Self-Distillation (MSD) and adaptive boosting ensemble segmenters. 
MSD conducts bidirectional distillation on the attention maps of the encoder and decoder, enhancing shallow-layer perception and refining deep-layer features. The adaptive boosting ensemble segmenter integrates sub-segmenters with different receptive fields, adapting to semantic features across a wider range of spatial frequencies. UMSPU expands the adaptable resolution range by 64 times, increasing it from 256×256 to 2048×2048. Moreover, it maintains a lightweight architecture and achieves a speed of 40 frames per second at high resolutions, propelling deep-learning phase unwrapping from research to practical level.

Our main work is summarized as follows:
\begin{enumerate}
   \item We propose a mutual self distillation (MSD) mechanism based on bidirectional feature refinement. MSD uses feature attention maps from corresponding encoder and decoder layers as mutual distillation targets, optimizing bidirectional Kullback-Leibler (KL) divergence. This provides cross-layer supervision ensuring stable semantic extraction across resolutions.
\item We construct an adaptive multi-scale ensemble segmentation architecture that integrates three sub- segmenters with different receptive fields. Through iterative weight optimization, we form a strong segmenter adaptable to a wider range of spatial frequencies, which reduces frequency bias and ensures semantic segmentation stability across various resolutions.

\item We propose a special curl loss. This loss is based on a convolution with fixed weights, which is used to estimate the curl in the predicted gradient field. Through this loss, the irrotationality of the gradient field output by the network is ensured.

\item Through experiments, We verify the adaptability of UMSPU at various resolutions, as well as its advantages in dealing with complex phase distributions, speed, and generalization. Furthermore, the practicality of UMSPU is also proved through experiments in structured light and InSAR.
\end{enumerate}

\begin{figure*}
    \centering
    \includegraphics[width = 2.0\columnwidth]{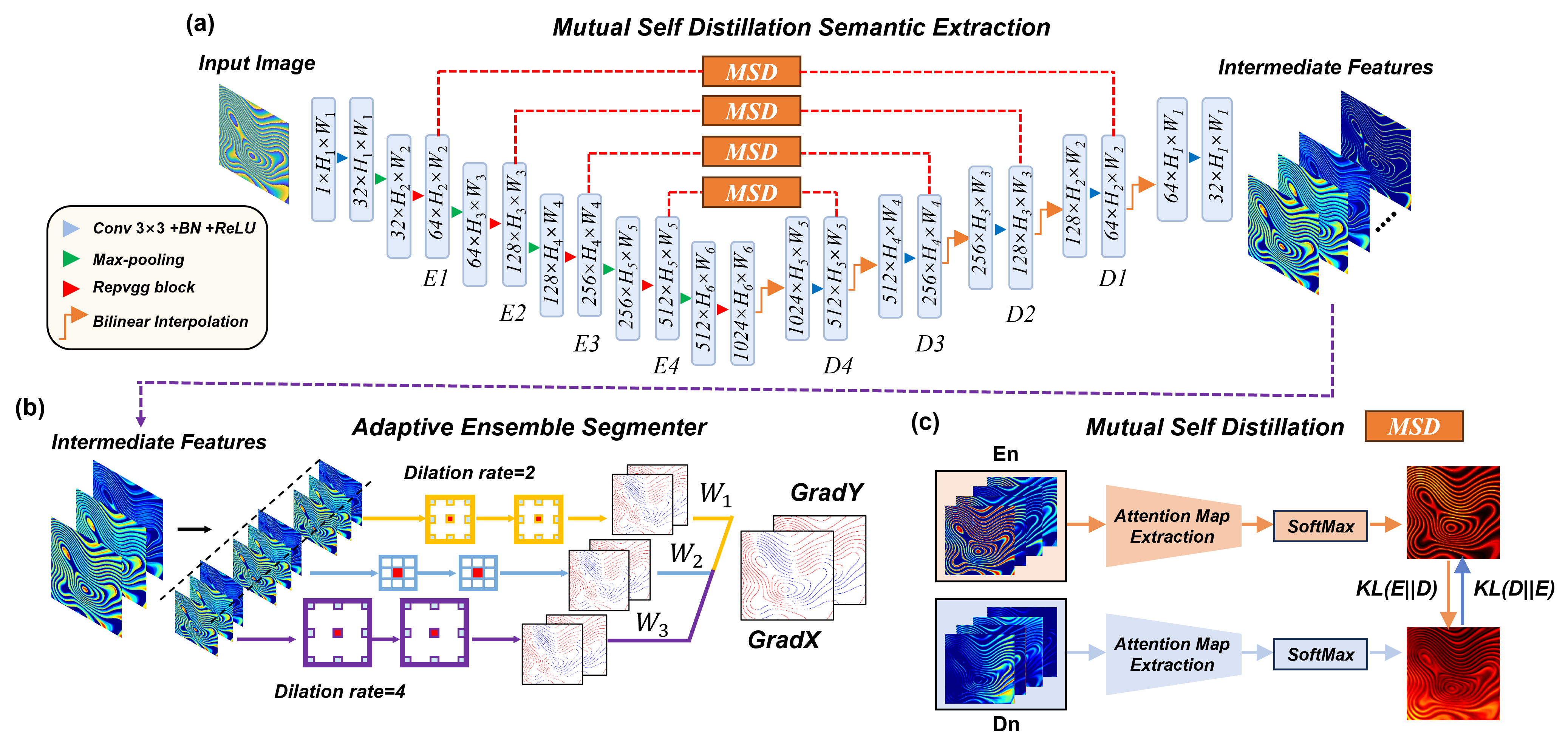}
    \caption{\rmfamily UMSPU comprises two components:
(a) Semantic Information Extraction via Mutual Self-Distillation: This mechanism leverages mutual self-distillation (MSD) to perform attention distillation on feature maps from the encoder and decoder at the same size, enabling mutual representation learning. As shown in (c), MSD extracts attention maps from the encoder and decoder feature maps, applies a softmax operation along the spatial dimension to generate attention soft labels, and computes bidirectional KL divergence for mutual distillation. (b) Adaptive Boosting Ensemble Segmenters: This component employs an adaptive boosting strategy to integrate three weak segmenters with different dilation rates into a strong segmenter, accommodating semantic features of varying spatial frequencies. The intermediate feature maps generated by (a) are passed to (b) to produce the final gradient segmentation result.}
    \label{fig:network}
     \vspace{-10pt} 
\end{figure*}

\section{Methodology}\label{}
\subsection{Overview}
The relationship between the unwrapped phase $\psi(x,y)$ and the wrapped phase $\varphi(x,y)$ can be described as follows:
\begin{equation}
    \label{phase}
    \psi(x,y)=\varphi(x,y)+2k(x,y)\pi 
\end{equation}
where $k(x,y)$ is the wrap count. We classify the wrap count gradient $\Delta k(x, y)$ of each point (-1, 0, +1) through a neural network, and then recover $k(x,y)$ via the discrete cosine transform to obtain the unwrapped phase. 

As shown in Fig. \ref{fig:network}, we propose the UMSPU to perform pixel-level wrap count gradient prediction. UMSPU consists of two components: (a) semantic information extraction and (b) ensemble segmenter. In (a), we propose the Mutual Self - Distillation (MSD) mechanism. This brand - new internal learning mechanism helps the network maintain stable semantic information extraction for inputs of various resolutions. In (b), through an adaptive boosting strategy, we perform weighted integration of multiple sub - segmenters, enabling it to have a wider range of spatial frequency adaptation. Additionally, we also design a curl loss to ensure that the gradient field output by UMSPU satisfies the physical constraint of irrotationality.

\subsection{Mutual Self-Distillation}
The wrap count gradient is classified into three categories: +1, 0, and -1. Among them, the points with a gradient of ±1 are very sparse across the entire image. This sparsity further intensifies as the image resolution increases, posing challenges to the encoder-decoder architecture. In low-size inputs, sparse semantic information diminishes through network layers, leading to blurred stripe features \cite{36}. In high-size inputs, capturing sparse semantic information requires broader feature perception, but shallow layers, constrained by small receptive fields, struggle to capture the long-range context necessary for locating stripe edges. This disrupts high-level feature propagation and hampers the recovery of sparse semantic cues, resulting in cumulative errors in stripe edge localization \cite{37}.
\par To enable the network to adapt to inputs of various resolutions, we propose a Mutual Self-Distillation (MSD) mechanism within the encoder-decoder architecture. The core principle of MSD is to extract the feature attention maps of the encoder and decoder during the network training process, and calculate the bidirectional Kullback - Leibler (KL) divergence. By optimizing the KL divergence loss, it enables complementary learning of the feature attention distributions between the encoder and the decoder. Decoder-to-encoder distillation helps the encoder capture the global context of these sparse features more effectively under high-resolution inputs, enabling the rapid localization of gradient distribution. Encoder-to-decoder distillation mitigates the degradation of sparse information in the deep layers under low-resolution inputs. 

The attention maps represent the regions of interest in each network layer and can be calculated from the feature maps as follows:
\begin{equation}
\label{attention_maps}
E_{\mathrm{sum}}^{(i,j)}=\sqrt{\sum_c(E^{(c,i,j)})^2},\enspace D_{\mathrm{sum}}^{(i,j)}=\sqrt{\sum_c(D^{(c,i,j)})^2}
\end{equation}
where $E^{(c,i,j)}$ and $D^{(c,i,j)}$ represent the values at $(i,j)$ in the $c$ channel of feature maps from encoder and decoder. Softmax function is then applied along the spatial dimension for normalization, generating the soft attention map labels:

\begin{equation}
\label{soft_labels1}
\begin{gathered}
E_{\mathrm{soft}}^{(i,j)}=\frac{\exp(E_{\mathrm{sum}}^{(i,j)})}{\sum\limits_{p=1}^{H}\sum\limits_{q=1}^{W}\exp(E_{\mathrm{sum}}^{(p,q)})} 
\end{gathered}
\end{equation}

\begin{equation}
\label{soft_labels2}
\begin{gathered}  
D_{\mathrm{soft}}^{(i,j)}=\frac{\exp(D_{\mathrm{sum}}^{(i,j)})}{\sum\limits_{p=1}^{H}\sum\limits_{q=1}^{W}\exp(D_{\mathrm{sum}}^{(p,q)})}  
\end{gathered}
\end{equation}
where $H$ and $W$ represent the height and width of the image. $E_{\mathrm{soft}}^{(i,j)}$ and $D_{\mathrm{soft}}^{(i,j)}$ are the values at $(i,j)$ in  soft attention map labels of encoder and decoder. Subsequently, the bidirectional KL divergence loss $KLloss$ is calculated as:
\begin{equation}
\label{KL_loss}
KLloss=\lambda_1\times KL\left(E\parallel D\right)+\lambda_2\times KL(D\parallel E) 
\end{equation}
\begin{equation}
\label{KL_ED}
\begin{gathered}
KL\left(E\parallel D\right)=\frac{1}{H\times W}\sum_{i=1}^{H}\sum_{j=1}^{W}E_{\text{soft}}^{(i,j)}\log\left(\frac{E_{\text{soft}}^{(i,j)}}{D_{\text{soft}}^{(i,j)}}\right) 
\end{gathered}    
\end{equation}
\begin{equation}
\label{KL_DE}
\begin{gathered}
KL\left(D\parallel E\right)=\frac{1}{H\times W}\sum_{i=1}^{H}\sum_{j=1}^{W}D_{\mathrm{soft}}^{(i,j)}\log\left(\frac{D_{\mathrm{soft}}^{(i,j)}}{E_{\mathrm{soft}}^{(i,j)}}\right) 
\end{gathered}    
\end{equation}
$KL\left(E\parallel D\right)$ denotes the KL divergence from the encoder's attention maps to the decoder's, where the encoder serves as teacher and the decoder as student. Conversely, $KL\left(D\parallel E\right)$ refers to the divergence in the opposite direction, with the decoder as teacher and the encoder as student. $\lambda_1$ and $\lambda_2$ are the weight coefficients for these two losses. $KL\left(E\parallel D\right)$ mitigates detail distortion in deep features for low-size inputs. $KL\left(D\parallel E\right)$ enhances shallow layers' contextual perception, enabling fine segmentation and dense prediction for high-size images. Therefore, during training, $KL\left(E\parallel D\right)$ is prioritized for low-size images and $KL\left(D\parallel E\right)$ for high-size ones. Accordingly, we set the weight coefficients $\lambda_1$ and $\lambda_2$ as follows:
\begin{equation}
\label{lambda1}
    \begin{gathered}
        \lambda_1=\frac{R_{\max}-R}{R_{\max}-R_{\min}} 
    \end{gathered}
\end{equation}
\begin{equation}
\label{lambda2}
    \begin{gathered}
        \lambda_2=\frac{R-R_{\min}}{R_{\max}-R_{\min}}
    \end{gathered}
\end{equation}
where $R$ represents the current image size, $R_{max}$ and $R_{min}$ correspond to the preset maximum and minimum sizes. As $R$ decreases, the proportion of $KL\left(E\parallel D\right)$ increases. As $R$ increases, the proportion of $KL\left(D\parallel E\right)$ increases.
\begin{figure*}[htbp]
    \centering
    \includegraphics[width = 2\columnwidth]{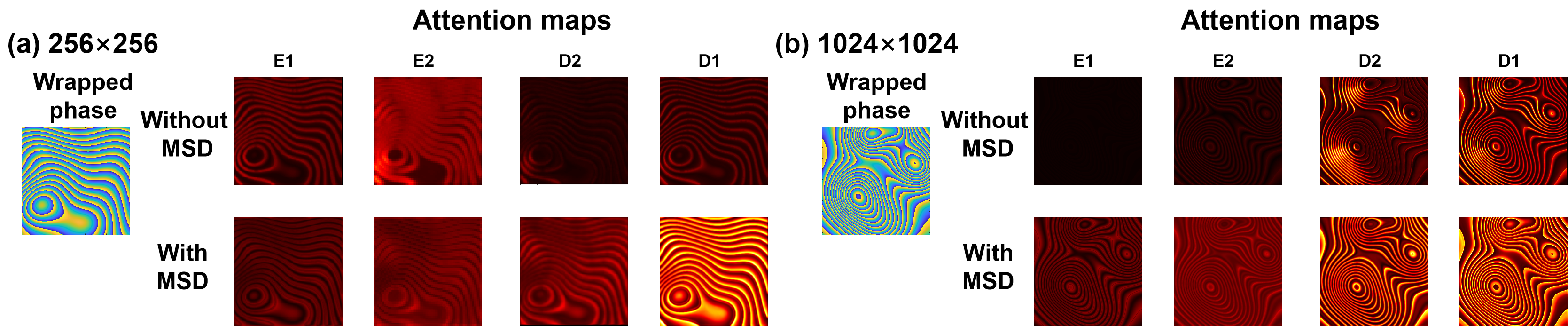}
    \caption{\rmfamily(a) Comparison of the attention maps of E1, E2, D2 and D1 before and after adding MSD with a 256×256 low-size input; (b) Comparison of the attention maps of E1, E2, D2 and D1 before and after adding MSD with a 1024×1024 high-size input.}
    \label{fig:attention-map}
     \vspace{-10pt} 
\end{figure*}

As shown in Fig. \ref{fig:network}(c), MSD connects the feature maps of the same size in the encoder and decoder, performs attention maps extraction and calculates two-way KL divergence losses for mutual attention distillation. It is worth mentioning that, to accelerate the network, we use the RepVGG Block as the basic block in the encoder. The RepVGG Block adopts the concept of structural re-parameterization, converting the multi-branch structure during training into a single-branch structure for inference \cite{38}. This significantly improves inference speed and reduces memory consumption. The detailed principles can be found in reference \cite{38}. 

As shown in Fig. \ref{fig:attention-map}, we compare the changes in the attention maps of the encoder E1, E2 and the decoder D1, D2 before and after introducing the MSD. Fig. \ref{fig:attention-map}(a) shows that after adding MSD, the deep layers of the network (D1, D2) focus more on the details at the phase cycle boundaries with low-resolution inputs of 256×256. This indicates that MSD helps the deep features retain critical details and enhances the recognition of stripes. Fig. \ref{fig:attention-map}(b) demonstrates that with high-resolution inputs of 1024×1024, the introduction of MSD allows the shallow parts of the network (E1, E2) to direct their focus towards the stripe regions. This suggests that MSD helps the network begin integrating semantic information at earlier stages, allowing it to identify important regions sooner. Additionally, the enhanced attention in E1 and E2 further increases the feature contrast in D2 and D1, improving the network's ability to distinguish between gradient and non-gradient points, which contributes to better segmentation accuracy in subsequent stages. Overall, MSD helps the network achieve more balanced and efficient semantic information extraction when processing inputs at different resolutions.

\subsection{Adaptive Boosting Ensemble Segmenters}
\label{section:Ensemble}
Another reason that restricts the network's adaptability to multi - resolution inputs is that, under different resolution inputs, there are differences in stripe density. Thus, the semantic information output by the decoder exhibits varying spatial frequencies. The single segmenter commonly placed at the end of the network is limited by its fixed convolutional kernel size, making it difficult to flexibly adapt to the various frequency components. This limitation causes the network to overly rely on and overfit to specific spatial frequencies, resulting in unstable segmentation performance with changing resolutions \cite{39}. To overcome this challenge, we integrate multiple sub-segmenters with different receptive fields, achieving comprehensive coverage and efficient utilization of multi-scale spatial frequency features. As shown in Fig. \ref{fig:network}(b), the ensemble segmenter consists of three distinct sub-segmenters, each using a 3×3 convolutional kernel as the base but with different dilation rates of 1, 2, and 4, respectively. This configuration provides the three sub-segmenters with different receptive fields, allowing them to focus on spatial frequency information at different scales.

\begin{figure}[]
    \centering
    \includegraphics[width = 1.0\columnwidth]{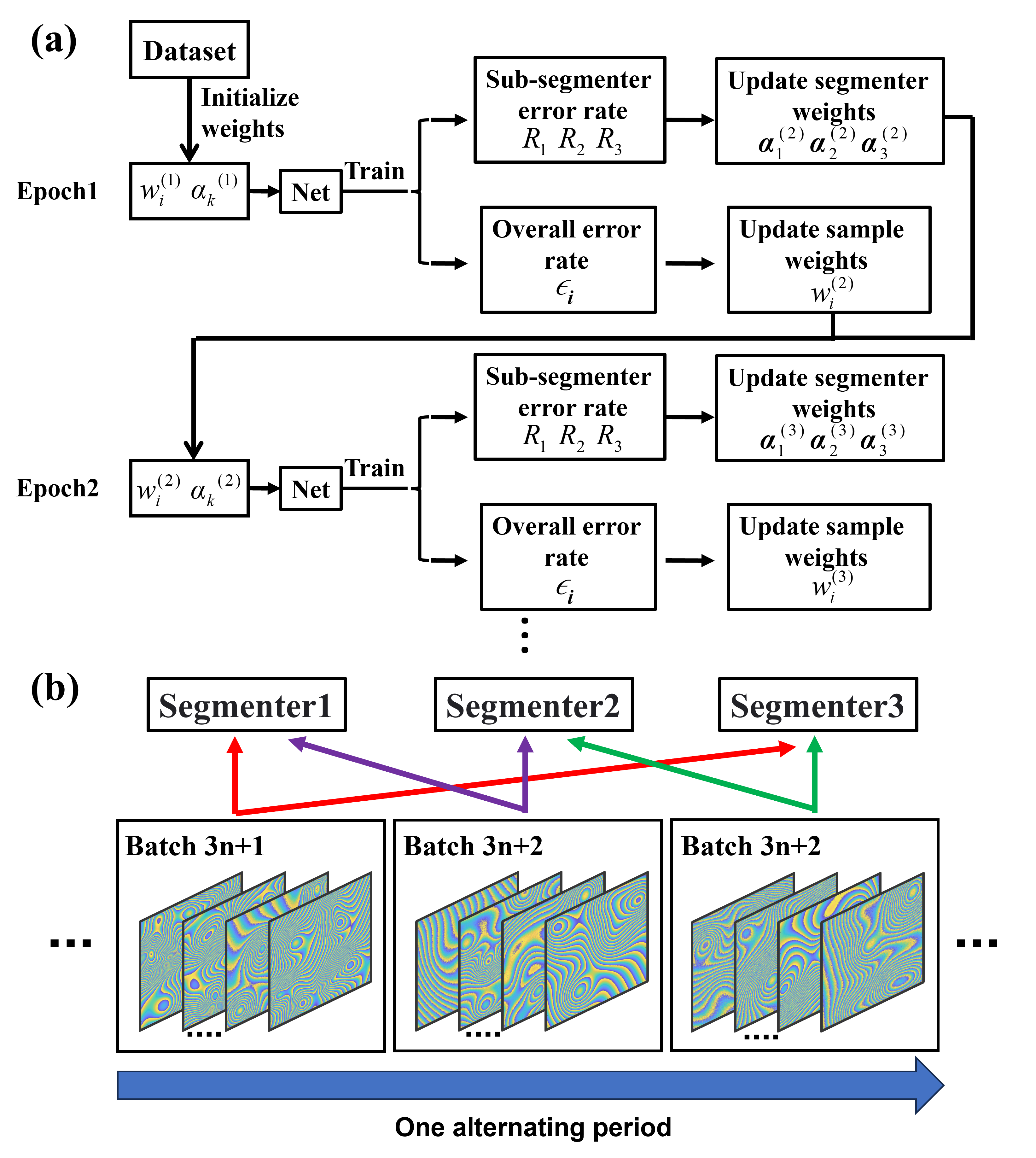}
    \caption{\rmfamily(a) In Adaptive Boosting, weak segmenter and sample weights are updated in every training round.(b) Three weak segmenters are paired into 3 training groups, with two sub-segmenters alternately selected for each training batch.}
    \label{fig:segmenter}
     \vspace{-10pt} 
\end{figure}

Under the ensemble learning framework, each weak segmenter dynamically adjusts its ability to capture frequency characteristics during the training process. However, due to the lack of clear prior information, it is difficult to assign appropriate weights to each segmenter before training. To ensure as much as possible that the multi-segmenter system achieves the widest and most complementary frequency coverage, we design an Adaboost algorithm that can dynamically update the weights of each segmenter.  As shown in Fig. \ref{fig:segmenter}, we design the specific training process as follows:

\noindent
\textbf{Step 1:} \textbf{Dataset Initialization and Cross-Training.} To avoid homogenization among the three sub-segmenters, we assign differentiated datasets during training. As shown in Fig. \ref{fig:segmenter}(b), We pair every two segmenters, resulting in three training pairs and alternately assign each input batch to a specific pair. This alternate mechanism ensures that each sub-segmenter is exposed to different combinations of sample sets, thereby enhancing its independence and diversity. During training, gradient accumulation is applied, with backpropagation performed every three batches. The encoder-decoder architecture is updated using the accumulated gradients, while each sub-segmenter is updated independently based on its respective gradients.\\
\noindent
\textbf{Step 2:} \textbf{Update the Weights of the Segmenters.} As shown in Fig. \ref{fig:segmenter}(a), between the adjacent epochs $\mathit{t}$ and $\mathit{t} + 1$, we update the weight of the $\mathit{k}$th segmenter from $\alpha_k^{(t)}$ to $\alpha_{k}^{(t+1)}$. This update is based on the error rate $R_{k}$ of the $\mathit{k}$-th segmenter on the dataset, where higher error rates correspond to smaller weights. Specifically, after each training round, we calculate $R_{k}$ as a weighted sum of individual sample errors:
\begin{equation}
        \label{RK}
          R_k = \sum_{i=1}^N w_i^{(t)} \epsilon_{i,k}
\end{equation}
\noindent where $N$ is the total number of samples, and $w_i^{(t)}$ is the weight of the $\mathit{i}$-th sample at epoch $t$, reflecting the difficulty of the sample. Initially, $w_i^{(0)} = 1/N$. The term $\epsilon_{i,k}$ denotes the error rate of the $\mathit{k}$-th segmenter on the $\mathit{i}$-th sample, calculated as:
\begin{equation}
        \label{epsilon_{i, k}}
           \epsilon_{i, k} = \frac{\sum\limits_j \gamma_{i, j} I(y_{i, j} \neq h_{k, i, j})}{\sum\limits_j \gamma_{i, j}}
\end{equation}
\noindent where $h_{k,i,j}$ is the prediction of the $\mathit{k}$-th segmenter for the $\mathit{j}$-th pixel of the $\mathit{i}$-th image, and $y_{i,j}$ is the corresponding ground truth label. The indicator function $I(y_{i,j} \neq h_{k,i,j})$ equals 1 if $y_{i,j} \neq h_{k,i,j}$, and 0 otherwise. For gradient mask points, $\gamma_{i,j} = 1$; for non-gradient points, $\gamma_{i,j} = 0$. Thus, only errors on gradient mask points are considered.

Finally, based on the error rate $R_{k}$, the weight of the $\mathit{k}$-th segmenter is updated using the following formula:
\begin{equation}
        \label{alpha_k^{(t+1)}}
           \alpha_k^{(t+1)} = \frac{\frac{1}{2}\ln\left(\frac{1-R_k}{R_k}\right)}{\sum\limits_{k=1}^{3} \frac{1}{2}\ln\left(\frac{1-R_k}{R_k}\right)}
\end{equation}
\textbf{Step 3:} \textbf{Sample Weights Update.} To emphasize samples with higher error rates in the next epoch, sample weights are updated accordingly. The pixel error rate $\epsilon_{i}$ of the $\mathit{i}$th image is calculated based on the weighted prediction results of the three segmenters:
\begin{equation}
        \label{epsilon_i}
        \setlength\abovedisplayskip{12pt}
        \setlength\belowdisplayskip{14pt}
        \epsilon_i = \frac{\sum\limits_j \gamma_{i,j} I\left(y_{i,j} \neq \hat{\nu}_{i,j}\right)}{\sum\limits_j \gamma_{i,j}}
\end{equation}
\noindent where $\mathit{y}_{i,j}$ represents the true label of the $\mathit{j}$th pixel in the $\mathit{i}$th image, and $\hat{\nu}_{i,j}$ is the aggregated result of the three segmenters. The indicator function $I(y_{i,j} \neq \hat{\nu}_{i,j})$ equals 1 if $y_{i,j} \neq \hat{\nu}_{i,j}$, and 0 otherwise. Only pixels on the gradient mask (where $\gamma_{i,j} = 1$) are considered.
Using the sample error rate $\epsilon_i$, the sample weights $w_i^{(t)}$ are updated as follows:
    \begin{equation}
        \label{$w_{i}^{(t+1)}$}
         \setlength\abovedisplayskip{12pt}
        \setlength\belowdisplayskip{14pt}
        w_i^{(t+1)} = \frac{w_i^{(t)}\exp\left(2\epsilon_i\right)}{\sum\limits_{i=1}^{N} w_i^{(t)}\exp\left(2\epsilon_i\right)}
    \end{equation}
\noindent where $w_i^{(t)}$ is the weight of the $\mathit{i}$th image at epoch $\mathit{t}$, and $w_i^{(t+1)}$ is the updated weight for the next epoch. Samples with updated weights $w_i^{(t+1)}$ below a threshold $\mathcal{I}$ are discarded, and new samples are introduced to enhance the model’s generalization. The weight for newly introduced samples is set to the mean weight $1/N$, and all sample weights are re-normalized before the next round of training.

By applying the above method to create a weighted ensemble of the three sub-segmenters, the network covers a broader frequency range, effectively avoiding the spatial frequency bias.

\subsection{Loss Function}
\label{section:Loss}
The loss function for wrap count gradient segmentation is designed to address two key challenges: (1) severe class imbalance, as the +1 and -1 classes have significantly fewer points compared to the 0 class, and (2) adherence to the physical constraint that gradient fields must be irrotational \cite{40}. To address these challenges, we propose a loss function that combines a weighted loss to handle class imbalance and a curl loss to enforce the irrotational constraint.

To mitigate the imbalance between classes, we design a weighted loss function, enabling the model to focus more on the segmentation accuracy of the +1 and -1 classes. The weighted loss consists of a weighted mean squared error (WMSE) loss and a weighted cross-entropy (WCE) loss, defined as:
\begin{equation}
\label{loss_equation1}
\begin{gathered}
    {L}_{wmse}=\frac1{C\times H\times W}\sum_i\sum_j\sum_c\beta_c(y_{c,i,j}-\hat{y}_{c,i,j})^2
\end{gathered}
\end{equation}
\begin{equation}
\label{loss_equation2}
\begin{gathered}
    {L}_{{wce}}=-\frac1{H\times W}\sum_i\sum_j\sum_c\beta_c\cdot y_{c,i,j}\cdot\log(\hat{y}_{c,i,j})
\end{gathered}
\end{equation}
where ${L}_{wmse}$ and ${L}_{{wce}}$ denote the WMSE and WCE losses, respectively. The one-hot encoded label at point $(i, j)$ in channel $c$ is $y_{c,i,j}$, and $\hat{y}_{c,i,j}$ is the corresponding model output after softmax normalization. The weight for the class corresponding to channel $c$ is denoted as $\beta_c$. Channel 0, 1, and 2 represent the 0, +1, and -1 classes, respectively. $C$, $H$, and $W$ denote the number of channels, height, and width of the image.

\begin{figure}[]
    \centering
    \includegraphics[width = 1.0\columnwidth]{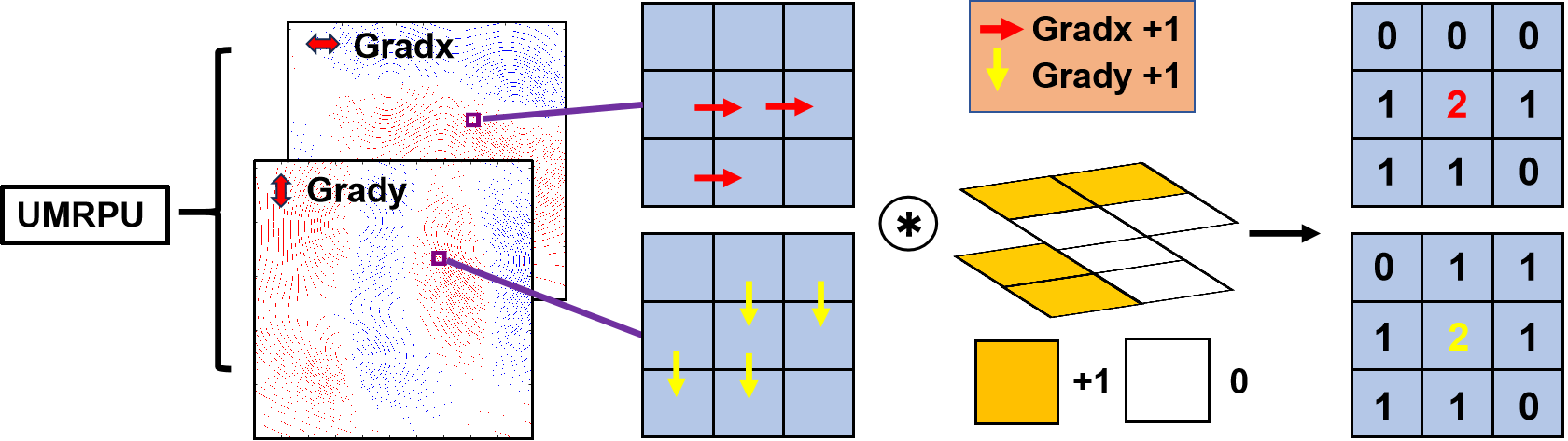}
    \caption{\rmfamily The curl estimation method uses two fixed convolution kernels to find curl points. After the convolution operation with these two fixed-weight convolutions, the points with values of 2 or -2 are identified as curl points.}
    \label{fig:derotation}
     \vspace{-10pt} 
\end{figure}

To ensure compliance with the irrotational constraint, we introduce a curl loss. This loss is based on a fixed-weight convolution-based curl estimation method. As shown in Fig. \ref{fig:derotation}, two fixed convolution kernels, $K_x$ and $K_y$, are applied to detect consecutive gradients in the horizontal and vertical directions:
\begin{equation}
\label{conv_kernels}
    K_x=\begin{pmatrix}1&1\\0&0\end{pmatrix}\quad K_y=\begin{pmatrix}1&0\\1&0\end{pmatrix}
\end{equation}

For gradx, the kernel $K_x$ detects consecutive gradients of ±1 along the horizontal direction, with the convolution result $f_x(i,j)$ being 2 or -2 at curl points. Similarly, for grady, the kernel $K_y$ detects vertical gradients. Curl points are identified as points with convolution results of 2 or -2. The curl loss is then defined as the ratio of curl points to gradient points:
\begin{equation}
\label{Lcurl}
L_{\text{curl}} = \frac{N_{\text{curl}}}{N_{\text{gradient}}}
\end{equation}
where \( N_{\text{curl}} \) and \( N_{\text{gradient}} \) represent the number of curl points and gradient points, respectively.

The loss function for each segmenter is defined as:\begin{equation}
\label{loss_seg}
    L_{sk}=L_{wmse}+L_{wce}+KLloss+L_{curl}
\end{equation}
where $L_{sk}$ is the loss for the $k$th segmenter, and $KLloss$ represents the KL divergence loss from MSD. The total loss for the model is defined as:
\begin{equation}
    \label{overall_loss}
    Loss=\alpha_1L_{s1}+\alpha_2L_{s2}+\alpha_3L_{s3}
\end{equation}
where $\alpha_1$, $\alpha_2$, and $\alpha_3$ are the weights assigned to segmenter 1, 2, and 3, respectively. By combining these losses, the model effectively balances class-specific accuracy, physical constraints, and multi-segmenter coordination, ensuring robust and consistent gradient segmentation.

\subsection{Phase Reconstruction}
\label{section:Reconstruction}
After obtaining the gradients in the x and y directions, we use the discrete cosine transform to reconstruct the wrap count gradient into the wrap count. Then, we multiply the wrap count by $2\pi$ and add the wrapped phase to obtain the final unwrapped phase. The detailed derivation is provided in the supplementary material.

\section{Experiments}
This section introduces our experimental framework. Section~\ref{Experimental Settings} details the experimental settings. Section~\ref{Comparison} compares UMSPU with other methods across different resolutions, different fringe densities, and processing speed. Section~\ref{Generalization} evaluates UMSPU’s stability and generalization under translational and rotational deformations. Finally, Section~\ref{Facial Morphology Reconstruction} and Section~\ref{Interferometric Synthetic Aperture Radar} demonstrate UMSPU’s practicality through comparisons in two real-world scenarios:  structured light three-dimensional reconstruction and Interferometric Synthetic Aperture Radar.
\subsection{Experimental Settings}
\label{Experimental Settings}
\subsubsection{Datasets}
Following the method in \cite{regression3}, we generate the unwrapped phase by superimposing multiple Gaussian distributions with different peaks and standard deviations onto different slope functions. Then, we calculate the wrapped phase and the wrap count. Different levels of noise are added to the wrapped phase, and the wrap count gradients in the x- and y-directions are obtained by taking the derivatives of the wrap count images. The wrapped phase image is the model input, while the x - and y -direction wrap count gradient images are the labels.

The dataset contains 12,000 samples, split into 80\% for training, 10\% for validation, and 10\% for testing. Sample resolutions range from 256×256 to 2048×2048, with SNRs between -2 dB and 4 dB. 
\subsubsection{Implementations}
The network is implemented using PyTorch, with training conducted on an NVIDIA A100 Tensor Core GPU and testing on an NVIDIA GeForce RTX 2060. The network is trained using the SGD optimizer with a batch size of 4, a learning rate of 1e-3, and a weight decay of 5e-4, reaching convergence after 300 epochs. During training, the sample weight threshold in \ref{section:Ensemble} is set to 5e-5. The class weights in \ref{section:Loss} are defined as [1, 10, 10], where class 0 is assigned a weight of 1, and classes 1 and -1 are assigned weights of 10.

In addition, we build a monocular structured light system. In this system, the optical module (S52 from Tengju Technology) is used to generate structured light, and the camera (MV - GE502GM from MindVision) is responsible for capturing the fringes. This system is used for the structured light experiments in the article.
\subsubsection{Evaluation Metrics}
Root mean square error (RMSE) is used to evaluate the phase unwrapping performance of the proposed and other methods. RMSE is defined as:
\begin{equation}
    \label{eval_loss}
    RMSE=\sqrt{\frac1{H\times W}\sum_{i=1}^H\sum_{j=1}^W\left(\hat{y}_{i,j}-y_{i,j}\right)^2}
\end{equation}

where $y_{i,j}$ represents the ground truth unwrapped phase at point $(i, j)$, $\hat{y}_{i,j}$ is the predicted unwrapped phase, $H$ and $W$ are the image height and width, respectively.
\subsection{Comparison}
\label{Comparison}
To validate the performance of UMSPU, we conduct comparative experiments with six commonly used phase unwrapping networks, including REDN \cite{count1}, PhaseNet \cite{count2}, PhaseNet2.0 \cite{count3}, DeepLabv3+ \cite{42}, VDENet \cite{regression3}, and GAUNet \cite{gradient3}. 
The comparisons include analyses under different resolutions, varying fringe densities, and model computational complexity.

\begin{figure*}[htbp]
    \centering
    \includegraphics[width=1.0\textwidth]{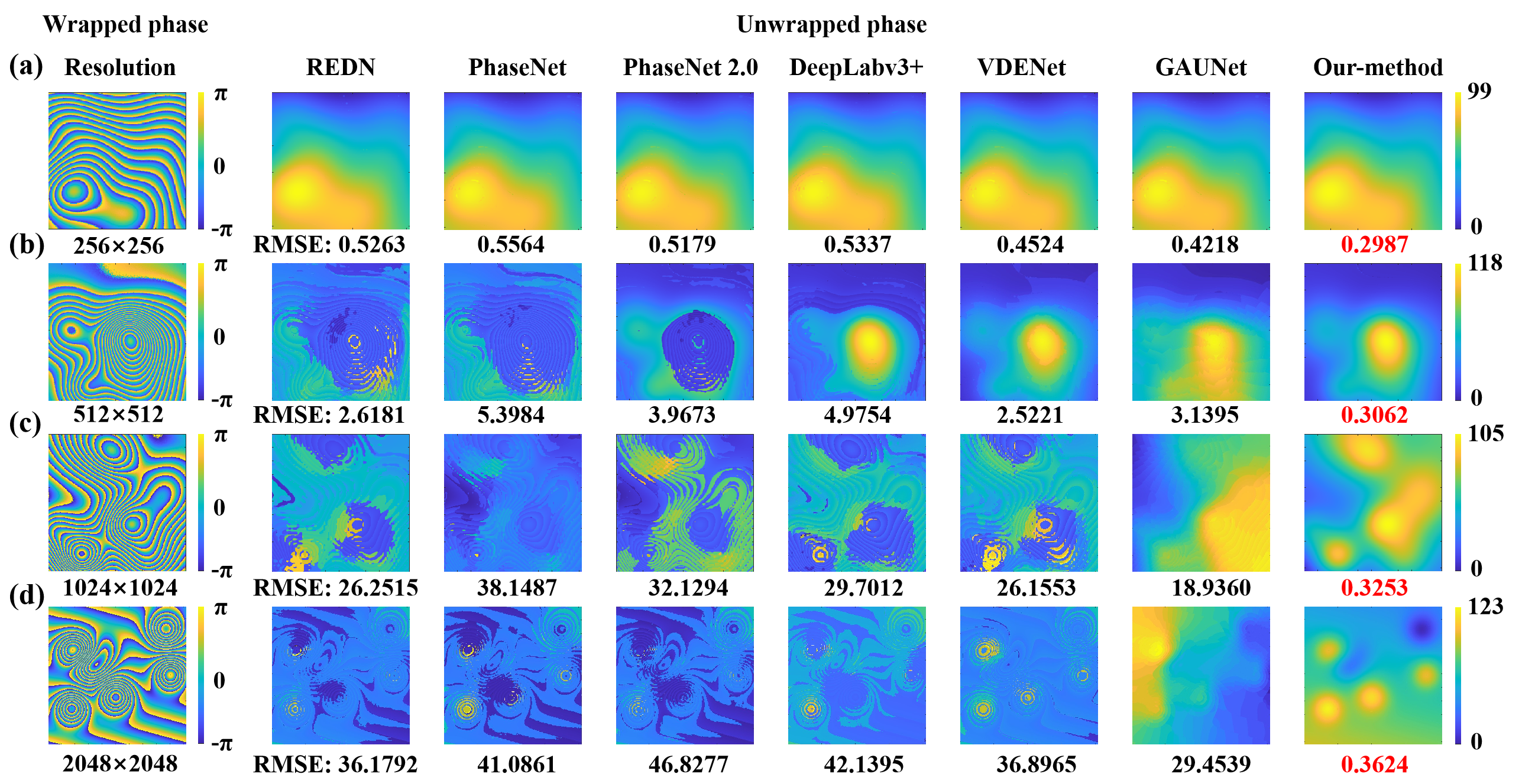}
    \caption{\rmfamily Comparison of the phase unwrapping performance and RMSE of REDN, PhaseNet, PhaseNet2.0, DeepLabv3+, VDENet, GAUNet and UMSPU at four resolutions: (a) 256×256, (b) 512×512, (c) 1024×1024 and (d) 2048×2048.}
    \label{fig:four-resolutions-performance}
     \vspace{-10pt} 
\end{figure*}

\begin{table*}\rmfamily
        \renewcommand\arraystretch{1.1}
	\caption{\rmfamily The mean RMSE of seven methods at different resolutions}
	\label{tableexample}
	\begin{center}
		\begin{tabular}{m{2cm}<{\centering} m{1.5cm}<{\centering} m{1.5cm}<{\centering} m{1.5cm}<{\centering} m{1.5cm}<{\centering} m{1.5cm}<{\centering} m{1.5cm}<{\centering}<{\centering} m{1.5cm}<{\centering}}
			\toprule  
			\multirow{2.4}{*}{Size} &\multicolumn{6}{c}{Method} \\ 
                \cmidrule{2-8}
                &REDN&PhaseNet&PhaseNet2.0&DeepLabv3+&VDENet&GAUNet& UMSPU \\
			\hline
	256$\times$256&0.5185&0.6269&0.5331&0.5427&0.4754&0.4322&\textbf{0.2954} \\
        512$\times$512&2.4752&5.5155&4.1424&4.8923&2.7527&3.3639&\textbf{0.3392} \\
    1024$\times$1024&24.9531&41.2977&33.2376&31.6656&25.5948&20.3161&\textbf{0.3429} \\
   2048$\times$2048 &37.2958&44.3165&42.2926&45.5863&39.8346&31.4632&\textbf{0.3483}\\
			\bottomrule
		\end{tabular}
	\end{center}
  \vspace{-10pt} 
\end{table*}

\subsubsection{Comparison at Different Resolutions}

We construct four test sets with resolutions of 256×256, 512×512, 1024×1024, and 2048×2048 to evaluate the performance of UMSPU and six other networks. As shown in Table \ref{tableexample}, all methods achieve low RMSE values at the lowest size (256×256). Among them, UMSPU performs best with a mean RMSE of 0.2954, followed by GAUNet, with a mean RMSE of 0.4322. This represents a 31.65\% improvement. As the resolution increases to 512×512, 1024×1024, and 2048×2048, UMSPU maintains high accuracy with mean RMSE values of 0.3392, 0.3429, and 0.3483, respectively. In contrast, the RMSE of the other six networks increases sharply and remains significantly higher than that of UMSPU.

To further illustrate this comparison, we randomly select one sample from each resolution for phase unwrapping using all methods. As shown in Fig. \ref{fig:four-resolutions-performance}, the other networks exhibit high accuracy only at the 256×256 resolution and fail at higher resolutions. Specifically, REDN, PhaseNet, and PhaseNet 2.0 (wrap count method), as well as DeepLabv3+ and VDENet (regression method), all produce RMSE values exceeding 35 at higher resolutions. This is because both wrap count and regression methods rely heavily on receptive field resolution. As input image resolution increases, the receptive fields of these networks fail to capture sufficient contextual information, leading to an inadequate understanding of details and the global structure in large images. GAUNet, as a gradient-based method, is less constrained by receptive field size but struggles with finer and more complex gradient variations at high resolutions due to its limitations in extracting complex structural features. This leads to error accumulation and gradient prediction distortion, resulting in RMSE values of 18.9360 and 29.4539 at 1024×1024 and 2048×2048, respectively. In contrast, UMSPU consistently achieves excellent phase unwrapping results across all resolutions, with RMSE values of 0.2987, 0.3062, 0.3253, and 0.3624 at the four resolutions, respectively. These results indicate that UMSPU overcomes the resolution limitations in phase unwrapping, showing excellent performance across various resolutions.

\subsubsection{Comparison at Different Densities}
To validate UMSPU's advantages in handling phases with different spatial frequencies, we construct four test sets for the experiment. The images in the test sets all have a size of 1024×1024 but differ in fringe density, with the numbers of fringes being 10($\pm3),30(\pm3),50(\pm3),and70(
\pm$3). We use UMSPU and six other methods on these four test sets and compare their mean RMSE. Notably, since UMSPU can handle high-resolution images, it directly performs phase unwrapping on the entire image. The other six networks are unable to effectively perform phase unwrapping on images with a resolution of 1024×1024 directly. Therefore, these six networks adopt a tiling strategy, where each image is divided into 16 regions of 256×256 pixels, processed individually, and then stitched back to the original size. During the stitching process, the gradients between different tiles are considered and accumulated to ensure continuity.

\begin{figure*}[h]
    \centering
    \includegraphics[width=1.0\textwidth]{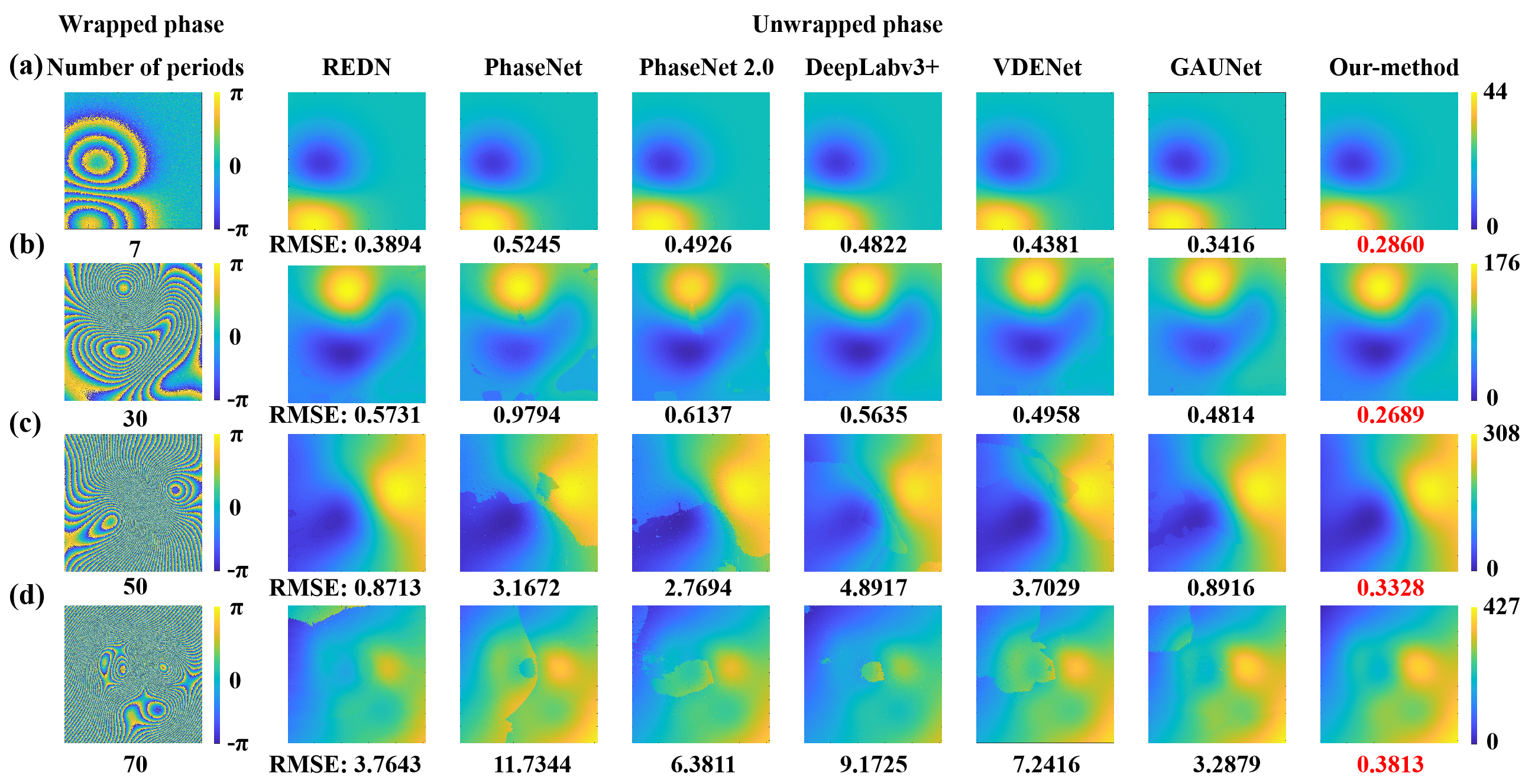}
    \caption{\rmfamily Phase unwrapping results of REDN, PhaseNet, PhaseNet 2.0, DeepLabv3+, VDENet, GAUNet and UMSPU at size of 1024×1024 under four fringe densities: (a) 7, (b) 30, (c) 50 and (d) 70. Notably, UMSPU directly predicts the entire large image, while the other six methods first split the image, perform block-wise predictions, and then stitch the results back together.}
    \label{fig:phase-cycle-densities}
     \vspace{-10pt} 
\end{figure*}
\begin{table}[t]\rmfamily
        \renewcommand\arraystretch{1.1}
	\caption{\rmfamily The mean RMSE of seven methods at different fringe Densities}
	\label{tableexample2}
	\begin{center}
		\begin{tabular}{m{1.5cm}<{\centering} m{1cm}<{\centering} m{1cm}<{\centering} m{1cm}<{\centering} m{1cm}<{\centering}}
			\toprule  
			\multirow{2.4}{*}{Method} &\multicolumn{4}{c}{Wrap count} \\ 
                \cmidrule{2-5}
                &10($\pm$ 3)&30($\pm$ 3)&50($\pm$ 3)&70($\pm$ 3)\\
			\hline
	REDN&0.3927&0.5669&0.9136&3.5423 \\
   PhaseNet&0.5477&0.6946&3.8121&13.2215\\
     PhaseNet2.0&0.4823&0.6370&2.4862&7.3429\\
        DeepLabv3+&0.4731&0.5576&5.3865&9.0873 \\
  VDENet&0.4239&0.5174&4.1882&7.6342\\
  GAUNet&0.3579&0.4634&0.8750&3.4766\\
 UMSPU&\textbf{0.2911}&\textbf{0.2957}&\textbf{0.3316}&\textbf{0.3728}\\
			\bottomrule
		\end{tabular}
	\end{center}
  \vspace{-10pt} 
\end{table}

As shown in Table \ref{tableexample2}, under low fringe densities of 10($\pm$3) and 30($\pm$3), all seven methods successfully perform phase unwrapping. The mean RMSEs of UMSPU are 0.2911 and 0.2957, respectively, while the second-best method, GAUNet, has mean RMSEs of 0.3579 and 0.4634, indicating that our method reduces RMSE by 18.66\% and 36.18\% compared to GAUNet. Even the worst-performing method, PhaseNet, maintains mean RMSEs of 0.5477 and 0.6946.

Under higher fringe densities of 50($\pm$3) and 70($\pm$3), UMSPU still achieves high-precision unwrapping, with mean RMSEs of 0.3316 and 0.3728, while the second-best GAUNet reaches 0.8750 and 3.4766, significantly higher than ours. Additionally, we randomly select one sample from each of the four densities. As shown in Fig. \ref{fig:phase-cycle-densities}, although the RMSE of UMSPU increases slightly with higher fringe densities, it remains as low as 0.3813 even at a fringe density of 70. This demonstrates that UMSPU adapts well to varying fringe densities.

In contrast, the other networks successfully unwrap phase at low fringe densities of 7 and 30, but fail at densities of 50 and 70. This highlights the superiority of UMSPU in handling phase distributions across various spatial frequencies.

\subsubsection{Model Computational Complexity}
\label{Model Computational Complexity}
We compare the computational complexity of UMSPU with six other networks under 1024×1024 resolution input. As shown in Table \ref{speed}, UMSPU has the lowest FLOPs, number of parameters, and inference time among all the networks. The running speed can reach over 40 frames per second under an ordinary graphics card. This demonstrates that UMSPU has significant advantages in terms of speed and lightweight design, making it easier to apply in real-world scenarios. Despite having the smallest computational complexity, UMSPU still achieves optimal performance, highlighting its efficiency. The efficiency comes from two aspects: first, the MSD mechanism enables the relatively simple network to achieve excellent performance through internal mutual representation learning during training. Second, we employ the RepVGG block as the basic block in the encoder, which uses structural reparameterization to convert the multi-branch topology during training into a single-path structure for inference, greatly optimizing the network's inference efficiency.
\begin{table}[h]\rmfamily
    \renewcommand\arraystretch{1.1}
    \caption{\rmfamily FLOPs, Params and Inference Time of Seven Networks at 1024×1024 Size}
    \label{speed}
    \begin{center}
        \begin{tabular}{m{1.5cm}<{\centering} m{1.5cm}<{\centering} m{1.5cm}<{\centering} m{1.5cm}<{\centering}}
            \toprule  
            \multirow{2}{*}{Method} & \multicolumn{1}{c}{FLOPS} & \multicolumn{1}{c}{Params} & \multicolumn{1}{c}{Inference Time} \\ 
            & (GFLOPs) & (million) & (ms) \\ \hline
            REDN & 951.30 & 17.71 & 756.14 \\
            PhaseNet & 495.93 & 13.40 & 321.07 \\
            PhaseNet2.0 & 68.22 & 7.97  & 80.25 \\
            DeepLabv3+ & 94.97 & 25.40 & 76.30 \\
            VDENet & 126.56 & 51.10 & 106.46 \\
            GAUNet & 770.83 & 31.04 & 472.32 \\
            UMSPU & \textbf{41.64} & \textbf{7.68} & \textbf{22.66} \\
            \bottomrule
        \end{tabular}
    \end{center}
     \vspace{-10pt} 
\end{table}

\subsection{Generalization}
\label{Generalization}

In practical phase imaging, data acquisition often involves varying positions and orientations, requiring the network to maintain consistent output under geometric transformations. Since training datasets cannot cover all possible data distributions, generalization is essential for model stability. To evaluate the generalization capability of the proposed method, we conduct experiments focusing on translation and rotation equivariance.

\begin{figure}[t]
    \centering
    \includegraphics[width = 0.95\columnwidth]{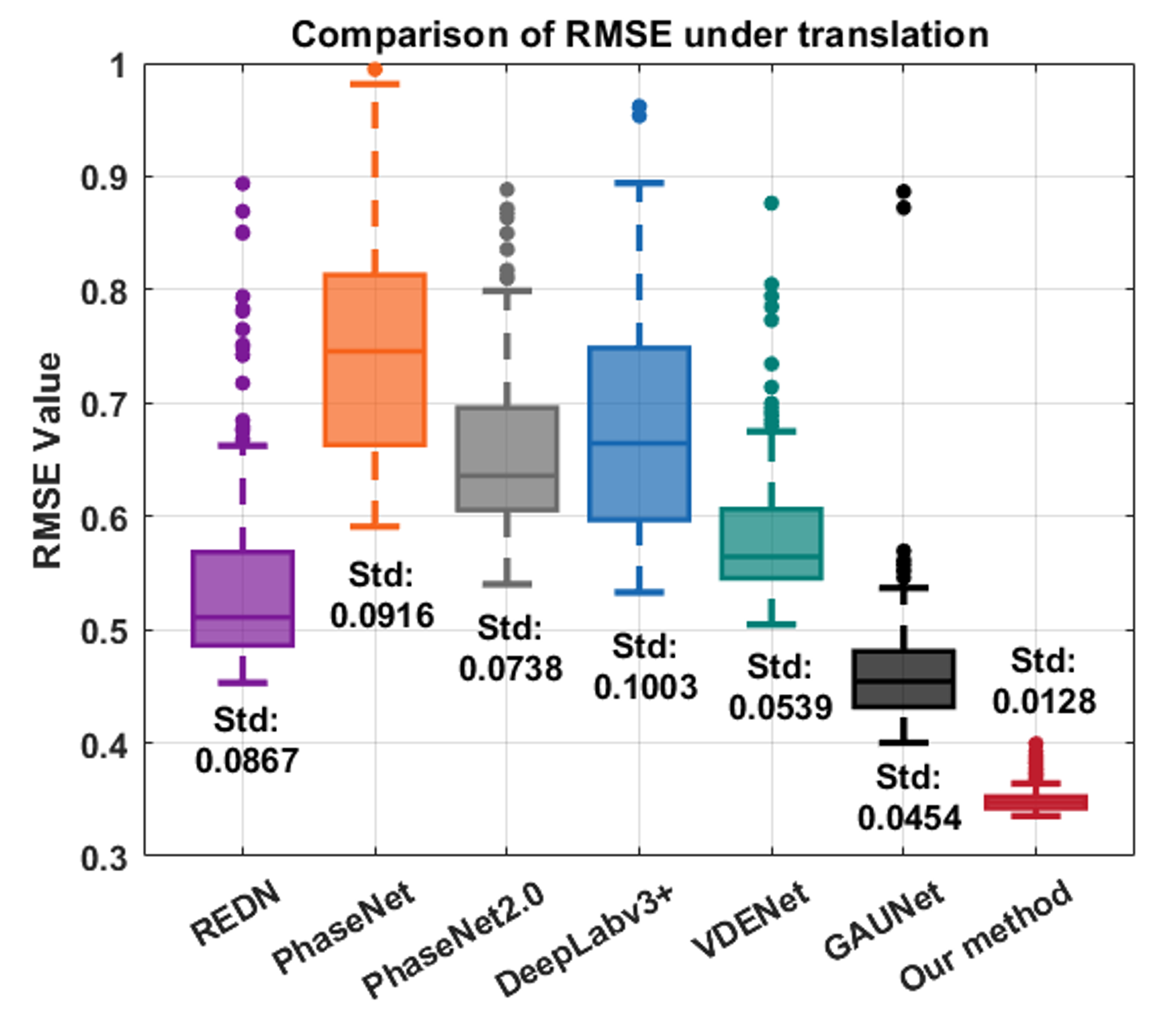}
    \caption{\rmfamily The box plot of the RMSE variation of seven methods during image translation.}
    \label{fig:translation-performance}
     \vspace{-10pt} 
\end{figure}

For translation equivariance, we use a 2000×2000 wrapped phase image with a 1024×1024 pixel anchor region. The anchor is translated across the image plane in 1-pixel steps, totaling 200 translations. Seven methods process the translated inputs, and the root mean square error (RMSE) is calculated after each translation. UMSPU directly processes the entire image, while the other six networks still adopt the strategy of block-wise prediction followed by stitching. Results are presented as box plots in Fig. \ref{fig:translation-performance}.

UMSPU achieves a mean RMSE of 0.3476 and median RMSE of 0.3404, significantly lower than the comparative methods, particularly GAUNet (mean RMSE 0.4618, median RMSE 0.4439). This demonstrates superior accuracy in handling translation transformations. In terms of stability, UMSPU exhibits a low RMSE standard deviation of 0.0128 and a peak-to-peak value of 0.0638, reflecting its robustness and minimal output variability under translations. In contrast, REDN, PhaseNet, PhaseNet 2.0, DeepLabv3+, VDENet, and GAUNet show much higher variability. Even GAUNet, the most stable one among the comparison methods, has a standard deviation of 0.0454, which is 3.55 times that of UMSPU, and a peak-to-peak value of 0.4865, which is 7.63 times that of UMSPU.

\begin{figure}[t]
    \centering
    \includegraphics[width = 0.95\columnwidth]{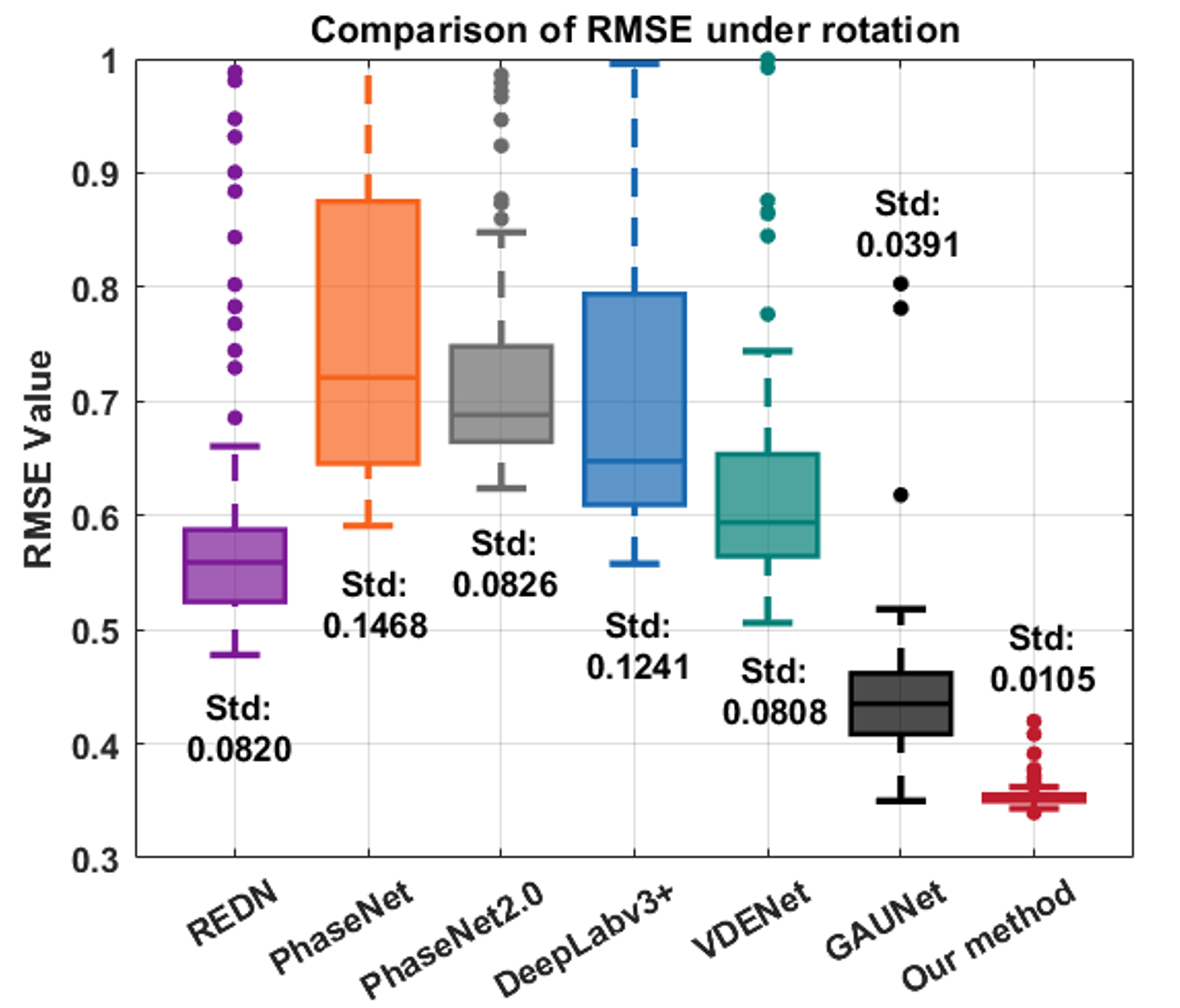}
    \caption{\rmfamily The box plot of the RMSE variation of seven methods during image rotation.}
    \label{fig:rotation-performance}
     \vspace{-10pt} 
\end{figure}

For rotation equivariance, we evaluate the model's performance on rotated inputs. A 1500-pixel diameter circular region is selected from a 2000×2000 wrapped phase image, with a 1024×1024 anchor box inside. During the experiment, the region rotates clockwise in 3° increments, causing corresponding rotational transformations within the anchor box. We analyze the RMSE distribution of seven methods in response to these rotations, with results shown in Fig. \ref{fig:rotation-performance}.

\begin{figure*}[htbp]
    \centering
    \includegraphics[width = 1.6\columnwidth]{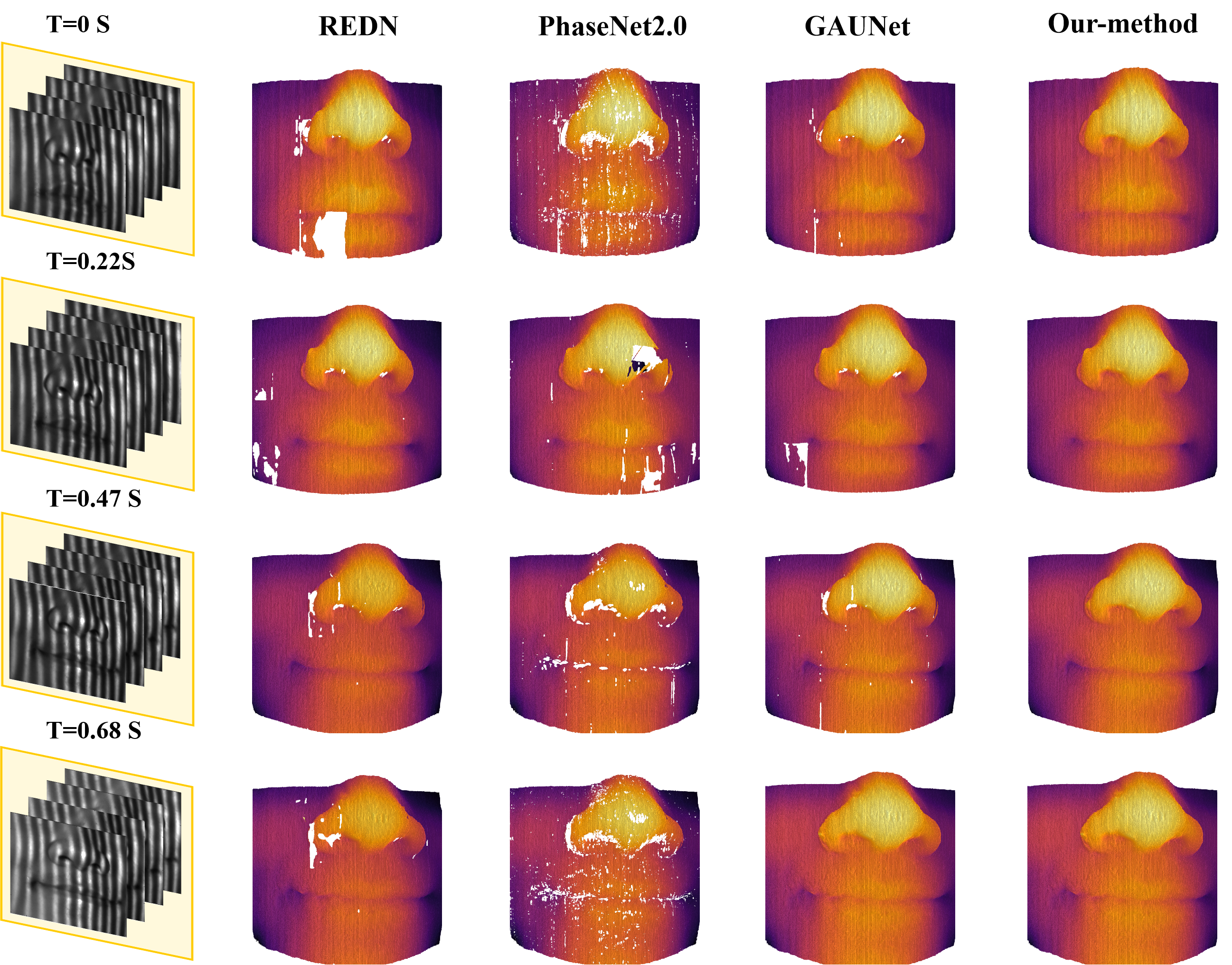}
    \caption{\rmfamily Facial morphology reconstruction results at four time points (T=0s, T=0.22s, T=0.47s, and T=0.68s) using REDN, PhaseNet2.0, GAUNet and UMSPU.}
    \label{fig:time}
     \vspace{-10pt} 
\end{figure*}
\begin{figure*}[b]
    \centering
    \includegraphics[width=0.85\textwidth]{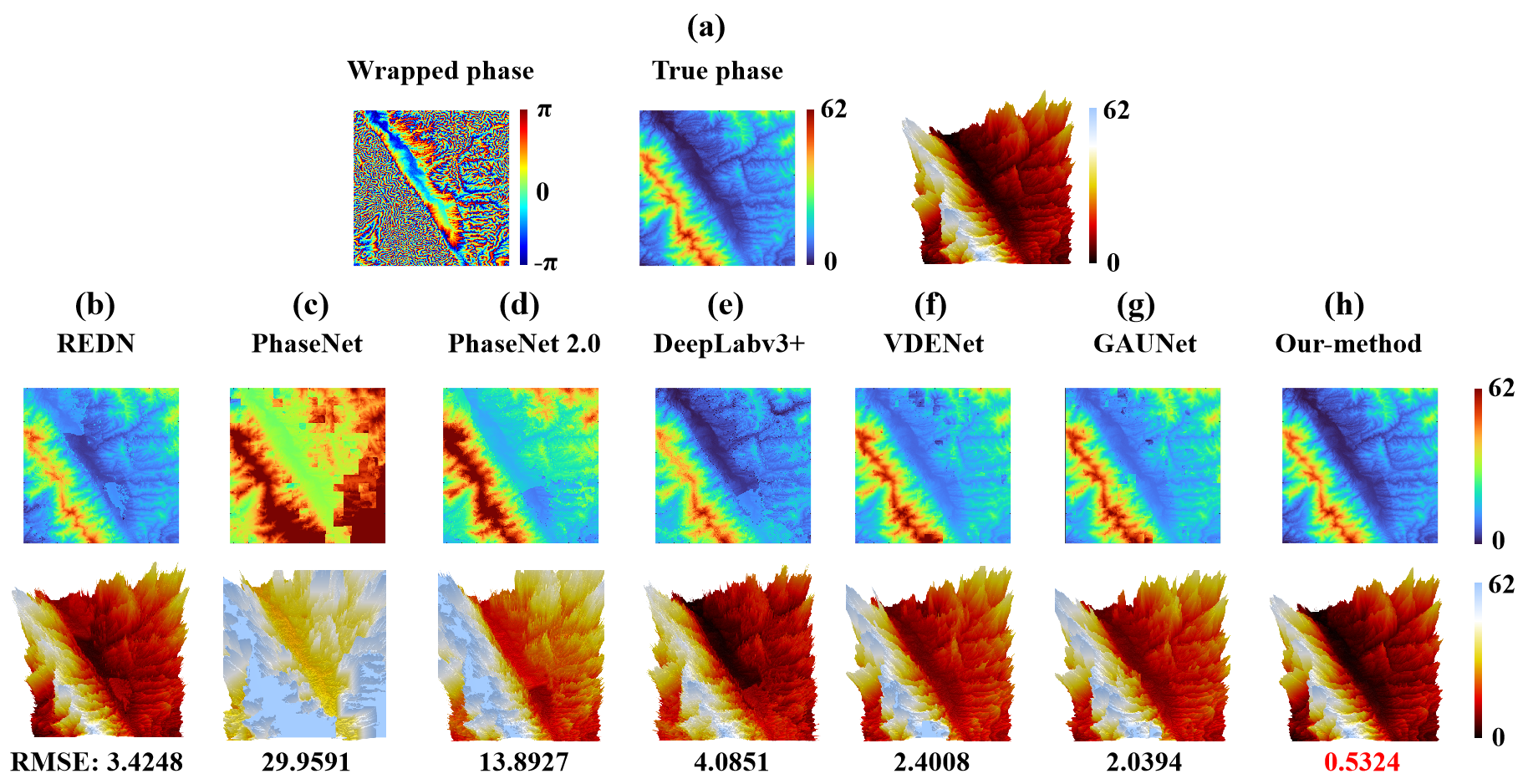}
    \caption{\rmfamily (a) The wrapped phase image generated from DEM data and the 2D and 3D representations of the unwrapped phase ground truth.(b)–(h) The 2D and 3D phase images computed by REDN, PhaseNet, PhaseNet 2.0, DeepLabv3+, VDENet, GAUNet and UMSPU, respectively.}
    \label{fig:yaogan}
     \vspace{-10pt} 
\end{figure*}
For rotational transformations, accuracy fluctuations for methods like REDN, PhaseNet, PhaseNet2.0, DeepLabv3+, VDENet, and GAUNet are more pronounced compared to the translation experiment. These methods show more outliers in their RMSE distributions, with some exceeding an RMSE value of 1, indicating difficulty in handling rotations. The RMSE standard deviations for these methods range from 0.0391 to 0.1468, reflecting varying degrees of instability. In contrast, UMSPU demonstrates exceptional stability, with an RMSE standard deviation of only 0.0105, significantly lower than the others. This indicates that UMSPU maintains high accuracy and consistency even with rotated inputs. Its robustness in handling these transformations further reinforces its superior performance in complex phase unwrapping tasks compared to existing methods.

\subsection{Structured Light Reconstruction Experiment}
\label{Facial Morphology Reconstruction}
 we also conduct facial morphology reconstruction experiments using the monocular structured light 3D system and obtain wrapped phase images using the four-step phase-shifting method. Then, we apply UMSPU and other methods for phase unwrapping to reconstruct the facial morphology. The resolution of the images is 2048×2048. UMSPU directly performs phase unwrapping on the entire image. Other methods adopt the stitching method because they are unable to handle high-resolution images. We display the reconstruction results of the four best-performing methods—REDN, PhaseNet2.0, GAUNet, and UMSPU—in Fig. \ref{fig:time}, In the experiment, we take the results obtained by the Gray code phase unwrapping method as the ground truth and calculate the differences between the results of the four methods and the ground truth. Pixels with the difference greater than $2\pi$ are set to NaN. 
 
Areas around the nostrils, and lips on the facial surface exhibit significant height variations, with complex fringe patterns and phase ambiguities. As shown, during the process of subtle facial expression changes, the results of REDN and PhaseNet2.0 in these regions exhibit obvious instability(Fig. \ref{fig:time}(a) and (b)). This indicates that the performance of REDN and PhaseNet2.0 in handling discontinuous phases is susceptible to disturbances from other regions. Although the results of GAUNet have better continuity, there are still certain errors at the boundaries of the phase cycles due to noise and high-frequency interference in the captured phase images (Fig. \ref{fig:time}(c)). In contrast, UMSPU (Fig. \ref{fig:time}(d)) achieves the smoothest and most stable results. It exhibits exceptional adaptability in handling discontinuous phase regions, accurately reconstructing complex morphologies in areas like the alar and nostrils. Even under challenging conditions with noise and high-frequency harmonic interference, UMSPU maintains high stability and accuracy. This proves the excellent stability and noise resistance of UMSPU.

\subsection{Interferometric Synthetic Aperture Radar}
\label{Interferometric Synthetic Aperture Radar}
To validate the practicality of UMSPU, we use data from the field of Interferometric Synthetic Aperture Radar (InSAR) to evaluate its performance under coherence conditions. A publicly available Digital Elevation Model (DEM) from TerraSAR-X/TanDEM-X, covering a specific area in Chongqing, is utilized to generate the corresponding unwrapped phase images and wrapped phase images, with both images having a resolution of 2048 × 2048.

As shown in Fig.\ref{fig:yaogan}, the selected region encompasses diverse terrains such as mountains, steep slopes, and valleys. The complexity of the phase distribution in this area is significantly higher than that in structured light measurement. Moreover, areas with dramatic topographical changes often involve issues like phase aliasing. Fig.\ref{fig:yaogan} illustrates a comparison of the results obtained by UMSPU and six other methods on this sample. It can be observed that the other six methods exhibit noticeable errors when dealing with such complex phase distribution images. Specifically, PhaseNet and PhaseNet 2.0 encounter global computational errors. The next-best method, GAUNet, shows gradient prediction confusion in areas with dense phase distributions, resulting in a phase unwrapping RMSE of 2.0394. In contrast, UMSPU maintains high accuracy on this sample, achieving an RMSE of only 0.5324. Notably, UMSPU does not incorporate data from the InSAR domain during training, and its success on this cross-domain sample highlights the strong robustness and generalization capabilities of UMSPU. These results highlight its technical superiority in addressing complex phase unwrapping challenges and underscore its broad applicability and practicality in real-world scenarios.

\section{Conclusion}

To address the issue that existing deep learning-based phase unwrapping networks lack the ability to perform phase unwrapping on high-resolution images, we propose a Universal Multi - Size Phase Unwrapping Network (UMSPU). This network can perform high - precision phase unwrapping on images within a 64 - fold resolution range, from 256×256 to 2048×2048, and maintain a phase unwrapping speed of dozens of frames per second even at high resolutions.

The superior performance of UMSPU mainly benefits from three factors. Firstly, we propose a Mutual Self-Distillation (MSD) mechanism. MSD conducts bidirectional attention distillation between the encoder and the decoder in the network, effectively preventing the loss of fine-grained features in the deep layers and enhancing the perception ability of the shallow layers. MSD enables effective cross-layer learning, endowing the network with the ability to extract fine-grained semantic information at different resolutions. Secondly, we propose an adaptive boosting ensemble segmenter. This segmenter, through an adaptive boosting strategy, integrates three sub-segmenters with different receptive fields in a weighted manner, enabling the network to cover a wider frequency range, effectively avoiding spatial frequency bias, and thus adapting to the differences in fringe density under different resolution inputs. Thirdly, we propose a curl loss, which ensures that the gradient field output by UMSPU satisfies the physical constraint of irrotationality. Thanks to the above three innovative factors, UMSPU, with a lightweight architecture, can achieve accurate phase unwrapping for inputs spanning a 64-fold resolution range.

To validate our method, we compare UMSPU with six other networks using both simulated and real-world data. The experimental results show that UMSPU significantly outperforms other networks under different resolutions and fringe densities, and exhibits excellent stability in translation and rotation experiments.
Meanwhile, through an experiment on face micro-expression reconstruction based on structured light and an experiment on synthetic aperture radar data, we effectively verify the practicality and generalization ability of UMSPU when confronted with complex phase distributions.

The outstanding performance of UMSPU in handling high-resolution images enables it to meet the strict high-resolution requirements of cameras in current practical measurement scenarios, thus producing detailed results. Its high-speed processing capabilities can satisfy the time-sensitive needs of real-time measurements, improving measurement efficiency. Moreover, the strong generalization and anti-interference capabilities of UMSPU provide a universal and reliable solution for phase measurements in various scenarios. For these reasons, UMSPU is expected to truly bring deep-learning-based phase unwrapping to the practical level.








\printcredits


\bibliographystyle{unsrt}
\bibliography{cas-refs}



\end{document}